\documentclass[conference,a4paper]{APSIPA2021}
\usepackage{graphicx}
\usepackage{amsmath}
\usepackage{amssymb}
\usepackage{amsfonts}
\usepackage{amsxtra}
\usepackage{mathrsfs}
\usepackage{latexsym}
\usepackage{multirow}
\usepackage{threeparttable}
\usepackage{url}
\usepackage{algorithm}
\usepackage[noend]{algpseudocode}
\usepackage{xcolor}
\usepackage{booktabs}
\usepackage{tabularx}
\usepackage{arydshln}
\usepackage{listings}
\usepackage{multirow}
\usepackage{tikz}
\usepackage{pgfplots}
\usetikzlibrary{arrows.meta,calc,decorations.markings,math,shapes.geometric,positioning}
\usepackage{siunitx}

\usepackage{mathtools}
\usepackage{cleveref}

\makeatletter
\let\MYcaption\@makecaption
\makeatother
\usepackage{subcaption}
\makeatletter
\let\@makecaption\MYcaption
\makeatother
\renewcommand{\vec}[1]{\boldsymbol{#1}}

\newcommand{\set}[1]{\mathbb{#1}}

\newcounter{num}

\DeclareMathOperator*{\argmax}{arg~max}

\definecolor{myorange}{rgb}{0.96, 0.67, 0}
\definecolor{mygreen}{rgb}{0.01, 0.69, 0.48}

\crefname{equation}{}{}
\crefname{figure}{Fig.}{Figs.}
\crefname{table}{Table}{Tables}
\crefname{algorithm}{Algorithm}{Algorithm}

\usepackage{geometry}
\usepackage{fancyhdr}

\fancypagestyle{firststyle}{
  \fancyhf{}
  \fancyhead[C]{2024 Asia Pacific Signal and Information Processing Association Annual Summit and Conference (APSIPA ASC)}
}

\pgfplotsset{compat=1.18}

\begin{document}

\title{Scene-Segmentation-Based Exposure Compensation for Tone Mapping of High Dynamic Range Scenes}

\author{
\authorblockN{
Yuma Kinoshita\authorrefmark{1} and
Hitoshi Kiya\authorrefmark{2}
}

\authorblockA{
\authorrefmark{1}
Tokai University, Japan\\
E-mail: ykinoshita@tokai.ac.jp}

\authorblockA{
\authorrefmark{2}
Tokyo Metropolitan University, Japan\\
E-mail: kiya@tmu.ac.jp}
}

\maketitle
\thispagestyle{firststyle}
\pagestyle{fancy}

\begin{abstract}
  We propose a novel scene-segmentation-based exposure compensation method
  for multi-exposure image fusion (MEF) based tone mapping.
  The aim of MEF-based tone mapping is
  to display high dynamic range (HDR) images on devices
  with limited dynamic range.
  To achieve this,
  this method generates a stack of differently exposed images
  from an input HDR image and fuses them into a single image.
  Our approach addresses the limitations of MEF-based tone mapping
  with existing segmentation-based exposure compensation,
  which often result in visually unappealing outcomes
  due to inappropriate exposure value selection.
  The proposed exposure compensation method first segments the input HDR image
  into subregions based on luminance values of pixels.
  It then determines exposure values for multi-exposure images
  to maximize contrast between regions
  while preserving relative luminance relationships.
  This approach contrasts with conventional methods
  that may invert luminance relationships or compromise contrast between regions.
  Additionally, we present an improved technique
  for calculating fusion weights to better reflect
  the effects of exposure compensation in the final fused image.
  In a simulation experiment to evaluate the quality of tone-mapped images,
  the MEF-based tone mapping with the proposed method outperforms
  three typical tone mapping methods including conventional MEF-based one,
  in terms of the tone mapped image quality index (TMQI).
\end{abstract}

\section{Introduction}
  High dynamic range (HDR) imaging, also known as HDR image synthesis,
  is a technique that reconstructs the luminance information
  of real-world scenes with an extensive dynamic range
  by combining multiple images captured
  using imaging sensors with a limited dynamic range
  ~\cite{debevec1997recovering,go2019image}.
  The resultant images are termed HDR images,
  which are distinctly categorized from images obtained
  through conventional photography, namely low dynamic range (LDR) images.
  In contemporary times,
  smartphones enables us to capture HDR images
  without clipped whites and crushed shadows.

  Dynamic range constraints are not limited to image sensors
  but also exist in display devices.
  The sRGB color space, which is widely used in current displays,
  assumes a relatively low peak luminance of \SI{80}{cd.m^{-2}}.
  While HDR displays capable of reproducing a broader dynamic range
  with peak luminances exceeding \SI{e3}{cd.m^{-2}} are becoming
  increasingly prevalent,
  the dynamic range of displays remains limited compared to real-world scenes.
  For instance, the luminance of direct sunlight can reach up to \SI{e9}{cd.m^{-2}}.
  To address this disparity and display HDR images on devices
  with restricted dynamic range, a process known as tone mapping is essential.
  This technique adjusts the image's tonal range to fit within
  the display's dynamic range capabilities
  ~\cite{reinhard2002photographic,durand2002fast,
  fattal2002gradient,drago2003adaptive,gu2013local,paris2015local}.

  Global tone mapping, which applies a single function
  such as a gamma curve or S-curve uniformly to all pixels,
  is a simple and widely used approach.
  However, this method tends to diminish the high local contrast
  between neighboring pixels inherent in HDR images.
  Consequently, research has been directed towards local tone mapping techniques,
  which adaptively apply tone mapping to individual pixels
  based on local information.
  Recently, there has been active research
  in multi-exposure image fusion (MEF) methods
  ~\cite{zokai2005image,mertens2009exposure,saleem2012image,wang2015exposure,
  li2014selectively,sakai2015hybrid,prabhakar2017deepfuse,kinoshita2019scene},
  which aim to represent the wide luminance range of real-world scenes
  within a limited dynamic range by fusing a stack of differently exposed images,
  called multi-exposure images.
  These fusion methods can be considered a variant of local tone mapping.

  While research on fusion algorithms has progressed,
  Kinoshita et al. demonstrated that
  the quality of the final fused image is significantly influenced
  by exposure values of input multi-exposure images
  ~\cite{kinoshita2018automatic_trans,kinoshita2019scene}.
  For this reason,
  they also proposed an exposure compensation method
  that segments the HDR image into subregions based on luminance
  and determines an exposure value for each subregion to ensure well-exposure.
  Although this method enables to avoid information loss in fused images
  due to clipped whites or crushed shadows,
  it does not necessarily contribute to obtaining a visually appealing image
  because of the inversion of luminance relationships
  and the loss of contrast between regions.

  Because of such a situation, in this paper,
  we propose an novel exposure compensation method
  for MEF-based tone mapping of HDR images.
  In this method, an input HDR image is first segmented into subregions
  based on luminance, similarly to the conventional method.
  Then, the exposure values of multi-exposure images are determined
  to maximize the contrast between regions.
  As a result, this method preserves
  the relative luminance relationships between regions,
  in contrast to the conventional method.
  Additionally,
  to better reflect the effects of exposure compensation in the final fused image,
  we also propose a method for calculating fusion weights.

  We evaluated the efficacy of the tone mapping scheme
  with the proposed segmentation-based exposure compensation method
  in terms of the quality of tone-mapped images by a simulation,
  where tone mapped image quality index (TMQI) was utilized as quality metrics.
  In the simulation,
  the tone mapping scheme with the proposed compensation method
  was compared with typical tone mapping methods, including a MEF based one.
  Experimental results showed that
  the proposed segmentation-based exposure compensation is effective
  for tone mapping.
  In addition, the segmentation-based exposure compensation outperformed
  three compared tone mapping methods in terms of TMQI.

\section{Related work}
  In this paper, we propose
  a novel scene-segmentation-based exposure compensation method
  for tone mapping HDR images.
  Because the proposed method is related to
  conventional tone mapping and multi-exposure image fusion,
  these techniques
  including a scene-segmentation-based exposure compensation method
  are summarized in this section.
  The following notations are used throughout this paper.
  \begin{itemize}
    \item Bold italic letters, e.g., $\vec{x}$, denote
      vectors or vector-valued functions,
      and they are assumed to be column vectors.
    \item The notation $\{x_1, x_2, \cdots, x_N\}$ denotes a set with $N$ elements.
      In situations where there is no ambiguity as to their elements,
      the simplified notation $\{x_n\}$ is used to denote the same set.
    \item The notation $p(x)$ denotes a probability density function of $x$.
    \item $U$ and $V$ are used to denote the width and the height
      of an input image, respectively.
    \item $\set{P}$ denotes the set of all pixels, namely,
      $\set{P} =
        \{(u, v)^\top | u \in \{1, 2, \cdots, U\} \land v \in \{1, 2, \cdots, V\}\}$.
    \item A pixel $\vec{p}$ is given as an element of $\set{P}$.
    \item An image is denoted by a vector-valued function
      $\vec{x}: \set{P} \to \set{R}^3$, where its output means
      RGB pixel values.
    \item The luminance of an image is denoted
      by a function $\ell: \set{P} \to \set{R}$.
  \end{itemize}

\subsection{Tone mapping}
  Display devices have limitations in their dynamic range,
  like imaging sensors.
  To present HDR images on such a display device,
  it is necessary to employ a tone mapping operator (TMO),
  which adjusts the luminance range of the images to fit
  within the dynamic range capabilities of the display
  ~\cite{reinhard2002photographic,durand2002fast,
  fattal2002gradient,drago2003adaptive,gu2013local,paris2015local}.

  The display luminance $\ell_d(p)$ of a tone-mapped image $\vec{y}$
  in the linear domain is generally calculated from
  the world luminance $\ell_w(p)$ of an input HDR image $\vec{E}$ as
  \begin{equation}
    \ell_d(p) = f(\ell_s(p)) = f(\ell_w (p) \Delta t),
    \label{eq:tonemapping}
  \end{equation}
  where $\Delta t > 0$ is a parameter that determines
  the brightness of the resulting image,
  and $f$ is a non-linear tone curve.
  This process relates to actual photography,
  where $\Delta t$ corresponds to the shutter speed that determines the exposure,
  and $f$ represents the camera response function that adjusts the tone.

  TMOs are primarily categorized into two types:
  global operators, which apply a common function $f$ to all pixels,
  and local operators, which apply different functions $f$
  to individual pixels based on local information.
  In the case of Reinhard's global operator~\cite{reinhard2002photographic},
  the parameter $\Delta t$ is given as
  \begin{equation}
    \Delta t = \frac{\alpha}{G(\ell_w | \set{P})}, \label{eq:reinahrd_ae} \\
  \end{equation}
  where $\alpha \in[0,1]$ is the parameter called “key value”,
  which indicates subjectively if the scene is light, normal, or dark
  and it is typically set as the middle gray $\ell_{\mathrm{gray}} = 0.18$
  of the display luminance on a scale from zero to one.
  The geometric mean $G(\ell_w | \set{P})$ of the world luminance $\ell_w(p)$
  is calculated as follows:
  \begin{equation}
    G(\ell_w | \set{P}) =
      \exp\left(\frac{1}{|\set{P}|}\sum_{p \in \set{P}} \log \ell_w(p)\right).
    \label{eq:geometric_mean}
  \end{equation}
  The TMO $f$ is given as
  \begin{equation}
    f(\ell) = \frac{\ell}{1 + \ell}\left(1 + \frac{\ell}{\ell_{\mathrm{white}}^2} \right)
    \: \mathrm{for} \: \ell \in [0, \infty),
    \label{eq:reinhardTMO}
  \end{equation}
  where parameter $\ell_{\mathrm{white}} > 0$ determines the white point.
  Specifically,
  for all $\ell$ where $\ell \geq \ell_{\mathrm{white}}$, $f(\ell) \geq 1$.
  The equality holds when $\ell = \ell_{\mathrm{white}}$.

  Although global TMOs are simple and widely used,
  it often results in a loss of local contrast inherent in HDR images.
  To address this issue, local tone mapping has gained attention.
  In particular, multiple exposure image fusion has become a focal point
  in this field.

\subsection{Multi-exposure image fusion}
  Multi-exposure image fusion (MEF) is a technique
  to represent the high dynamic range of a scene as an LDR image.
  MEF combines several images
  of the same scene captured at different exposure levels by varying shutter speeds
  \cite{zokai2005image,mertens2009exposure,saleem2012image,wang2015exposure,
  li2014selectively,sakai2015hybrid, nejati2017fast, ma2017robust}.
  A stack of differently exposed images is referred to as ``multi-exposure images.''
  Given a stack of $M$ exposure images $\vec{x}_1, \cdots, \vec{x}_M$,
  a fused image $\vec{y}$ is produced:
  \begin{equation}
    \vec{y} = \mathscr{F}(
      \vec{x}_1, \cdots, \vec{x}_M),
    \label{eq:fusion}
  \end{equation}
  where $\mathscr{F}$ indicates a function for fusing $M$ images
  into a single image.

  MEF methods can be used to tone-map HDR images
  by performing exposure compensation for a given HDR image
  to generate multi-exposure images, which are then combined.
  This exposure compensation is done by
  substituting $M$ distinct shutter speeds $\Delta t_1, \cdots, \Delta t_M$
  into \cref{eq:tonemapping}.
  For this reason, MEF methods can be
  considered as a type of local tone mapping methods.

  To produce high-quality fused images from given differently exposed images
  while preserving local contrast of original scenes,
  many research works have been focused on the development of the fusion method $\mathscr{F}$.
  Mertens et al.\cite{mertens2009exposure} proposed a multi-scale fusion scheme in which
  contrast, color saturation, and well-exposedness measures are used
  for computing fusion weights.
  In the work by Nejati et al. \cite{nejati2017fast},
  base-detail decomposition is applied to each input image,
  and base and detail layers are then combined individually.
  An exposure fusion method based on sparse representation,
  which combines sparse coefficients with fused images,
  was also proposed in \cite{wang2015exposure}.

  While research on fusion algorithm has progressed,
  Kinoshita et al. demonstrated that
  the quality of the final fused image is significantly influenced
  by not only fusion algorithm
  but also exposure values or shutter speeds $\Delta t_1, \cdots, \Delta t_M$
  used for generating multi-exposure images $\vec{x}_1, \cdots, \vec{x}_M$.
  Furthermore, they proposed a scene-segmentation-based exposure compensation method
  to automatically determine the exposure of input images,
  ensuring that the fused image does not contain over- or under-exposed regions.

  The key idea of this method is to segment the scene into $M$ small regions
  based on luminance values
  by clustering them using a Gaussian mixture model (GMM).
  In the segmentation,
  the distribution of luminance $\ell_w$ is modeled with $M$ Gaussians as
  \begin{equation}
    p(\log{\ell_w}) = \sum_{m=1}^{M} \pi_m \mathcal{N}(\log{\ell_w} \ | \ \mu_m, \sigma_m)
    \label{eq:gmm}
  \end{equation}
  and a pixel $\vec{p} \in \set{P}$ is assigned to a subset $\set{P}_m$ of $\set{P}$
  when $m$ satisfies
  \begin{equation}
    m = \argmax_k
      \frac{
        \pi_k \mathcal{N}(\log{\ell (\vec{p})} \ | \ \mu_k, \sigma_k)
      }{
        \sum_{j=1}^{M} \pi_j \mathcal{N}(\log{\ell (\vec{p})} \ | \ \mu_j, \sigma_j)
      }.
    \label{eq:clustering}
  \end{equation}
  Given $\set{P}_m$,
  the exposure of the $m$th image $\vec{x}_m$ is determined
  so that the $m$th region $\set{P}_m$ achieves proper exposure.

\subsection{Scenario}
\begin{figure}[!t]
    \centering
    \begin{subfigure}[t]{\hsize}
      \centering
      \includegraphics[width=\columnwidth]{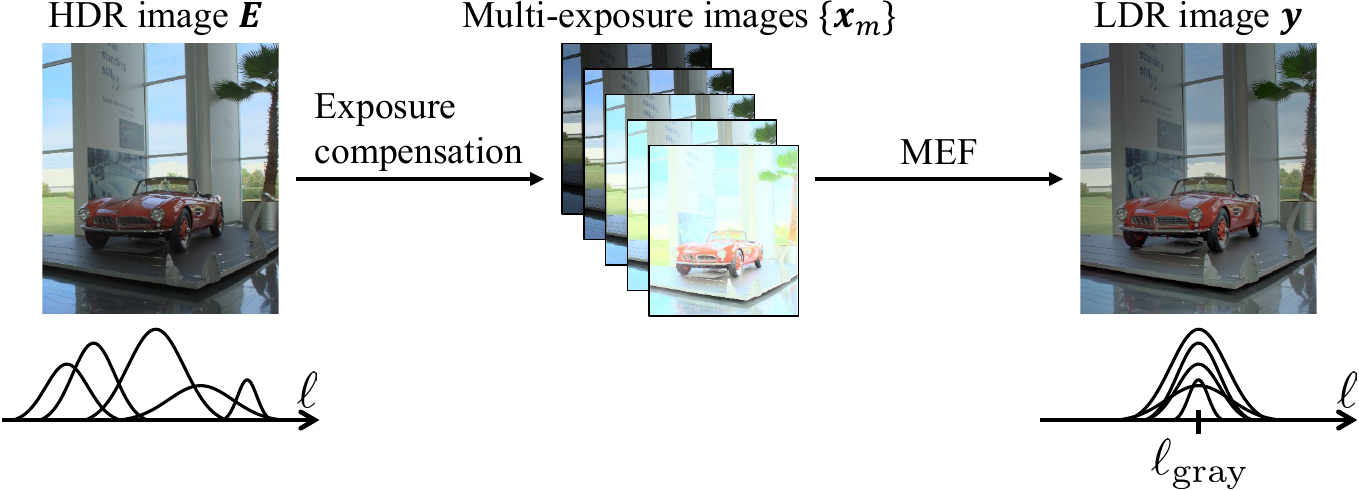}
      \caption{Conventional \label{fig:conv}}
    \end{subfigure}\\
    \begin{subfigure}[t]{\hsize}
      \centering
      \includegraphics[width=\columnwidth]{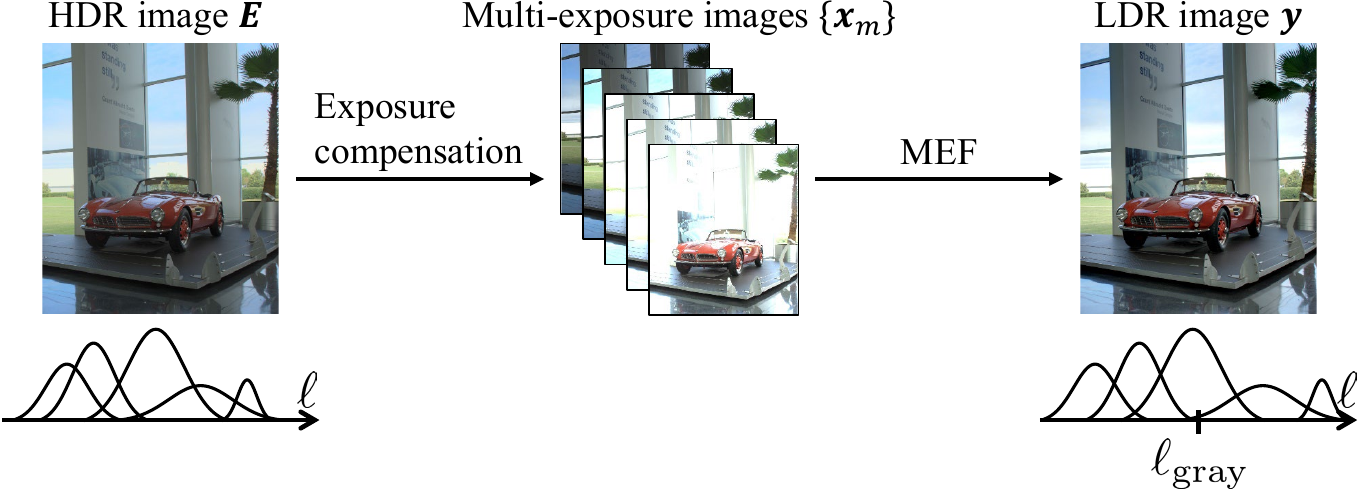}
      \caption{Proposed \label{fig:prop}}
    \end{subfigure}
    \caption{Conventional and proposed exposure compensation based on scene segmentation}
  \end{figure}

  In this paper, we focus on the exposure values of multi-exposure images
  for MEF, like the aforementioned literatures
  ~\cite{kinoshita2018automatic_trans,kinoshita2019scene}.
  In the literatures,
  $\Delta t_m$ is calculated so that the geometric mean of luminance
  on $m$th region $\set{P}_m$
  equals the middle-gray $\ell_{\mathrm{gray}}$, as
  \begin{equation}
    \Delta t_m = \frac{\ell_{\mathrm{gray}}}{G(\ell_w | \set{P}_{m})}.
    \label{eq:unknownEV}
  \end{equation}
  By determining exposures in this manner,
  it is possible to avoid information loss due to highlights or shadows
  because many MEF methods assign greater weight
  to pixels having luminance values close to middle gray during the fusion process.
  However, the resulting fused image tends to have the average luminance
  of each region $\set{P}_m$ converging towards the middle gray (see \cref{fig:conv}).
  This can lead to issues such as the inversion of luminance relationships
  and the loss of contrast between regions.

  Because of such a situation, in this paper,
  we propose an novel exposure compensation method that preserves
  the relative luminance relationships between regions, for MEF-based tone mapping.
  Additionally,
  we also propose a method for calculating fusion weights
  to better reflect the effects of exposure compensation in the final fused image.

\section{Proposed tone mapping scheme}
  The goal of the proposed method is to represent
  a HDR scene with a wide dynamic range as an LDR image
  while preserving the relative luminance relationships and local contrast.
  To achieve this,
  this study first divides the scene into multiple regions,
  similar to conventional approaches.
  Subsequently,
  the exposure of multi-exposure images is determined
  to maximize the contrast between regions (see \cref{fig:prop}).
  HDR images are subjected to exposure compensation
  to generate multi-exposure images,
  which are then composited to obtain the final image.
  During the compositing process,
  to reflect the effects of exposure compensation,
  the weights used in the composition are determined
  based on the exposure value and the luminance value of each pixel.

\subsection{Exposure compensation}
  In this method, we consider a relative exposure value (EV)
  with respect to $\ell_{\mathrm{gray}}$ as a reference point.
  In accordance with \cref{eq:tonemapping,eq:reinahrd_ae},
  we normalize the world luminance $\ell_w$ to 0 EV using the following equation:
  \begin{equation}
    \ell_s (\vec{p}) = \frac{\ell_{\mathrm{gray}}}{G(\ell_w | \set{P})} \ell_w (\vec{p}).
    \label{eq:normalize_luminance}
  \end{equation}
  The distribution of the scaled luminance $\ell_s$ is modeled as
  a Gaussian mixture model (GMM)
  and $\set{P}$ is divided into $M$ regions $\{\set{P}_m\}$,
  similar to \cref{eq:gmm,eq:clustering}:
  \begin{align}
    &p(\log{\ell_s}) = \sum_{m=1}^{M} \pi_m \mathcal{N}(\log{\ell_s} \ | \ \mu_m, \sigma_m),
      \label{eq:prop_gmm}\\
    &m = \argmax_k
      \frac{
        \pi_k \mathcal{N}(\log{\ell_s (\vec{p})} \ | \ \mu_k, \sigma_k)
      }{
        \sum_{j=1}^{M} \pi_j \mathcal{N}(\log{\ell_s (\vec{p})} \ | \ \mu_j, \sigma_j)
      }
    \label{eq:prop_clustering}
  \end{align}
  where we assume that mean values $\{ \mu_m \}$ is sorted in ascending order
  and thus $\set{P}_1$ and $\set{P}_M$ indicates the darkest and the brightest regions,
  respectively.
  In the regions $\{ \set{P}_m \}$,
  a reference region $\set{P}_{m_{\mathrm{ref}}}$ is selected
  based on the mixture component that contributes most significantly
  to $\ell_{\mathrm{gray}}$.
  This selection is mathematically defined as:
  \begin{equation}
    m_{\mathrm{ref}}
    = \argmax_k \pi_k \mathcal{N}(\log{\ell_{\mathrm{gray}}} \ | \ \mu_k, \sigma_k)
    \label{eq:prop_ref}
  \end{equation}

  After the segmentation,
  we perform the exposure compensation considering
  the relative luminance relationships between regions.
  The exposure compensation factors $\{ \Delta t_m \}$ are determined
  as the ratio of the target mean $\mu'_m$ to
  the mean $\mu_m$ of the $m$-th region:
  \begin{equation}
    \Delta t_m = \frac{\exp(\mu'_m)}{\exp(\mu_m)}.
    \label{eq:prop_exposure}
  \end{equation}
  Considering the dynamic range of the display luminance,
  we fix the target mean of the darkest region and the brightest region
  to $v_{\mathrm{min}}$ EV and to $v_{\mathrm{max}}$ EV, respectively:
  \begin{equation}
    \mu'_{1} = \log{2^{v_{\mathrm{min}}} l_{\mathrm{gray}}}, \quad
    \mu'_{M} = \log{2^{v_{\mathrm{max}}} l_{\mathrm{gray}}}.
    \label{eq:prop_edge_target}
  \end{equation}
  To maximize contrast between regions,
  we calculate the target mean evenly spaced in the log-luminance domain:
  \begin{equation}
    \mu'_m =
    \begin{dcases}
      \frac{\mu_{m_{\mathrm{ref}}} - \mu'_1}{m_{\mathrm{ref}}-1}(m-1) + \mu'_1 & \ 1 < m < m_{\mathrm{ref}} \\
      \mu_{m_{\mathrm{ref}}} & \ m = m_{\mathrm{ref}} \\
      \frac{\mu'_M - \mu_{m_{\mathrm{ref}}}}{M-m_{\mathrm{ref}}}(m-m_{\mathrm{ref}}) + \mu_{m_{\mathrm{ref}}} & \ m_{\mathrm{ref}} < m < M
    \end{dcases}.
    \label{eq:prop_target}
  \end{equation}

  Finally,
  we apply the exposure compensation and tone mapping
  to obtain the luminance of the $m$th exposure image $\vec{x}_m$ as:
  \begin{equation}
    \ell_m (\vec{p}) = f(\ell_s (\vec{p}) \Delta t_m),
    \label{eq:prop_compensation}
  \end{equation}
  where we use Reinhard's global TMO shown in \cref{eq:reinhardTMO}
  as the tone curve $f$
  with $\ell_{\mathrm{white}} = 2^{v_{\mathrm{white}}} \ell_{\mathrm{gray}}$.

\subsection{Weight calculation for multi-exposure image fusion}
  In our approach,
  we employ Mertens' well-knowm MEF method~\cite{mertens2009exposure}
  as the fusion function $\mathscr{F}$.
  This fusion is performed by averaging
  the Laplacian pyramid decomposition of input multi-exposure images $\{ \vec{x}_m \}$ as
  \begin{equation}
    \vec{L}_l[\vec{y}](\vec{p})
      = \sum_{m=1}^{M} G_l[w_m](\vec{p}) \vec{L}_l[\vec{x}_m](\vec{p}),
    \label{eq:mertens_fusion}
  \end{equation}
  where $\vec{L}_l[\vec{x}_m](\vec{p})$ is the $l$th level of the Laplacian pyramid
  of the $m$th image $\vec{x}_m$ at pixel $\vec{p}$,
  and $G_l[w_m](\vec{p})$ is the $l$th level of the Gaussian pyramid
  of the fusion weight $w_m$ for the $m$th exposure image at pixel $\vec{p}$.

  To reflect the effect of the exposure compensation in the fused image,
  this method employs novel fusion weights.
  To calculate the weight $w_m$ for the $m$th exposure image $\vec{x}_m$,
  we define the difference between the luminance $\ell_m$
  and the target mean $\mu_m'$ as
  \begin{equation}
    d_m(\vec{p}) = g(\ell_m(\vec{p})) - g(\exp(\mu_m')),
    \label{eq:prop_diff}
  \end{equation}
  where $g(\ell) = (\gamma \circ f) (\ell)$ is a composite function 
  of a gamma correction function $\gamma$ and the tone curve $f$
  because the fusion in \cref{eq:mertens_fusion} is performed
  in the gamma-corrected sRGB domain.
  Using the difference $d_m(\vec{p})$,
  the weight $w_m$ is calculated as
  \begin{equation}
    w_m(\vec{p})  = \frac{
        \exp(-(d_m(\vec{p}))^2)
      }{
        \sum_{k=1}^M \exp(-(d_k(\vec{p}))^2)
      }.
    \label{eq:prop_weight}
  \end{equation}
  This weighting scheme emphasizes pixels whose luminance is close
  to the target mean $\mu_m'$ of their corresponding region $\set{P}_m$.

\subsection{Proposed procedure}
  The procedure of MEF-based tone mapping from an input HDR image $\vec{E}$
  into $\vec{y}$,
  with the proposed exposure compensation and the fusion weight calculation,
  is summarized as follows:
  \begin{enumerate}
    \item Exposure compensation:
      \begin{enumerate}
        \item Calculate the world luminance $\ell_w$ of the input image $\vec{E}$.
        \item Normalize the world luminance $\ell_w$ to 0 EV
          by using \cref{eq:normalize_luminance}.
        \item Separate $\set{P}$ into $M$ areas $\{\set{P}_m\}$
          by using \cref{eq:prop_gmm,eq:prop_clustering}
          to segment image $x$ into $M$ areas.
        \item Determine the target mean $\mu'_m$ for each region $\set{P}_m$
          by using \cref{eq:prop_ref,eq:prop_edge_target,eq:prop_target}.
        \item Calculate the exposure compensation factor $\Delta t_m$
          for each region $\set{P}_m$
          by using \cref{eq:prop_exposure}.
        \item Tone map the input image $\vec{E}$ according to
          \cref{eq:prop_compensation}.
        \item Generate multi-exposure images $\{\vec{x}_m\}$ in the sRGB color space as
          \begin{equation}
            \vec{x}_m = \vec{\gamma} \left( \frac{\ell_m(\vec{p})}{\ell_w(\vec{p})} \vec{E}(\vec{p}) \right),
          \end{equation}
          where $\vec{\gamma}$ is a function that applies gamma correction
          to each of the RGB values.
      \end{enumerate}
    \item Multi-exposure fusion:
      \begin{enumerate}
        \item Calculate fusion weights $\{ w_m \}$ by using \cref{eq:prop_diff,eq:prop_weight}
          from $\{ \ell_m \}$ and $\{ \mu'_m \}$.
        \item Fuse the Laplacian pyramid of the multi-exposure images $\{ \vec{x}_m \}$
          into that of the fused image $\vec{y}$ by using \cref{eq:mertens_fusion}.
        \item Obtain an tone-mapped image $\vec{y}$
          by reconstructing the fused Laplacian pyramid $\vec{L}_l[\vec{y}]$.
      \end{enumerate}
  \end{enumerate}

\section{Simulation}
  We evaluated the tone mapping scheme
  that uses the proposed segmentation-based exposure compensation
  and the proposed fusion weight,
  in terms of the quality of tone-mapped images $\vec{y}$.
  Hereinafter, the scheme is simply called the proposed method.

\subsection{Simulation conditions}
  In the simulation,
  20 HDR images in the literature~\cite{korshunov2015subjective}
  were used as input images $\vec{x}$.
  Each image $\vec{x}$ was tone-mapped by the proposed method
  and the quality of the resulting image $\vec{y}$ was evaluated.

  The following four tone-mapping methods were compared in this paper:
  Reinhard's global tone mapping (Global)~\cite{reinhard2002photographic},
  Reinhard's local tone mapping (Local)~\cite{reinhard2002photographic},
  MEF with conventional exposure compensation (Conventional)~\cite{kinoshita2018automatic_trans,kinoshita2019scene},
  and the proposed method (Proposed).
  For the two MEF based methods (Conventional and Proposed),
  Mertens' method~\cite{mertens2009exposure} was used as
  a fusion function $\mathscr{F}$.
  In this simulation, we set the parameters in the proposed method as
  $v_{\mathrm{min}} = -3, v_{\mathrm{max}} = 1.5$, and $v_{\mathrm{white}} = 2.5$.

  To evaluate the quality of tone-mapped images by each of the four methods,
  objective quality assessments are needed.
  Typical quality assessments such as the peak signal to noise ratio (PSNR)
  and the structural similarity index (SSIM) are not suitable
  for tone-mapped images
  because they needs a reference tone-mapped image with the highest quality.
  We therefore used the tone mapped image quality index (TMQI)
  ~\cite{yeganeh2013objective}.

\subsection{Results}
\begin{figure}[!t]
    \centering
    \begin{subfigure}[t]{0.45\hsize}
      \centering
      \includegraphics[width=\columnwidth]{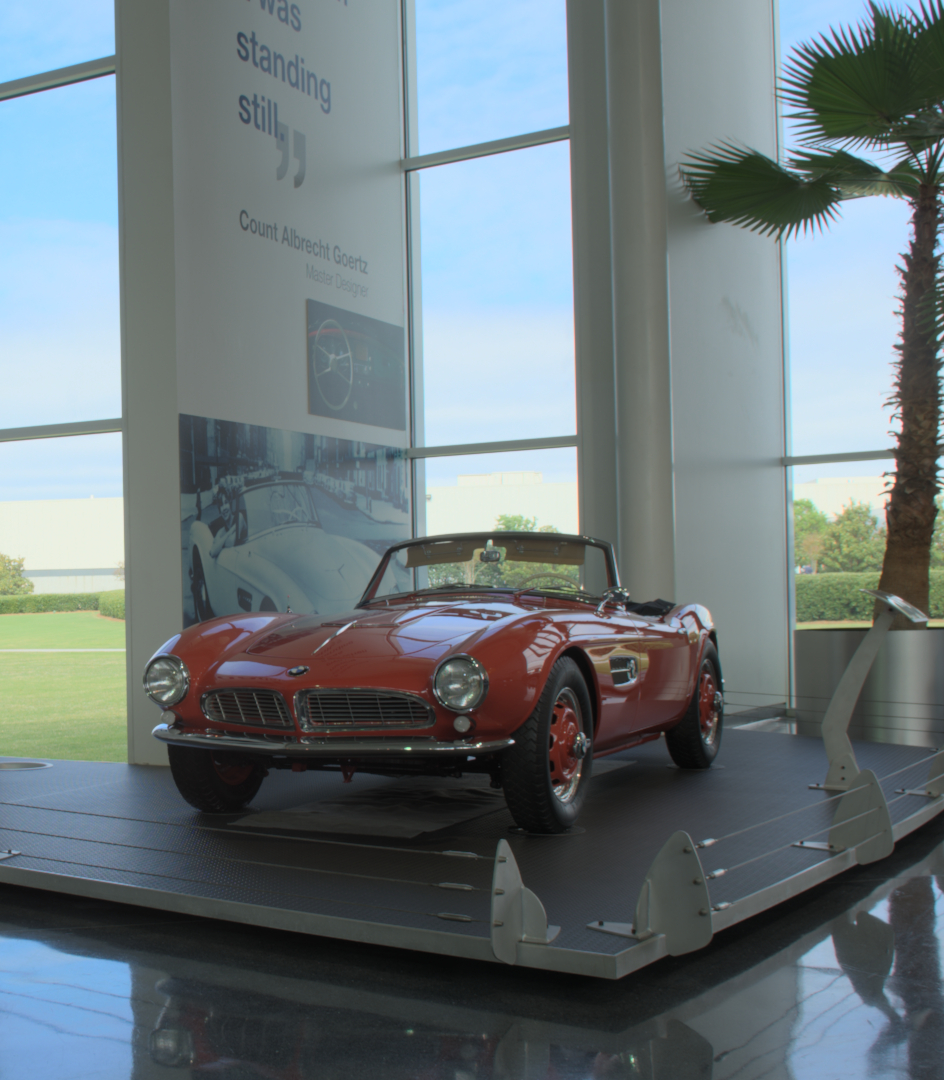}
      \caption{Global~\cite{reinhard2002photographic}\label{fig:507_global}}
    \end{subfigure}
    \begin{subfigure}[t]{0.45\hsize}
      \centering
      \includegraphics[width=\columnwidth]{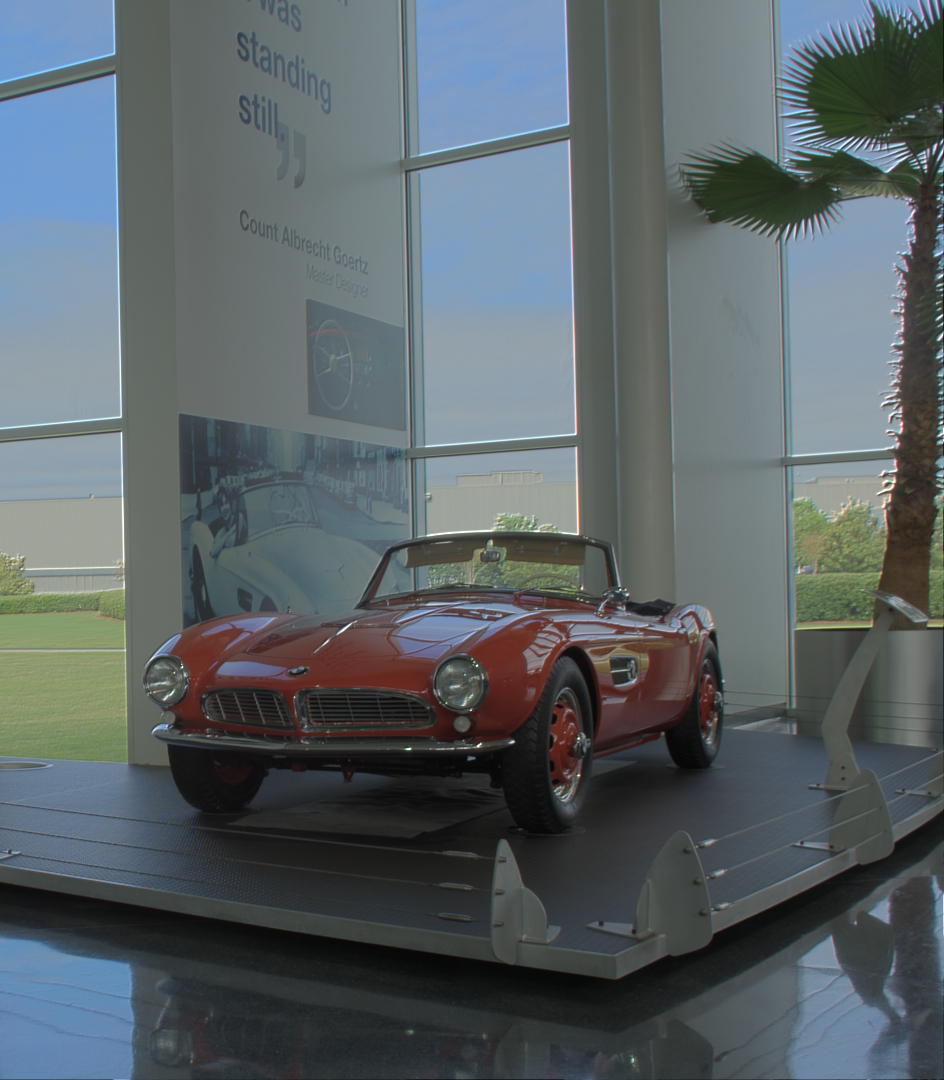}
      \caption{Local~\cite{reinhard2002photographic}\label{fig:507_local}}
    \end{subfigure}\\
    \begin{subfigure}[t]{0.45\hsize}
      \centering
      \includegraphics[width=\columnwidth]{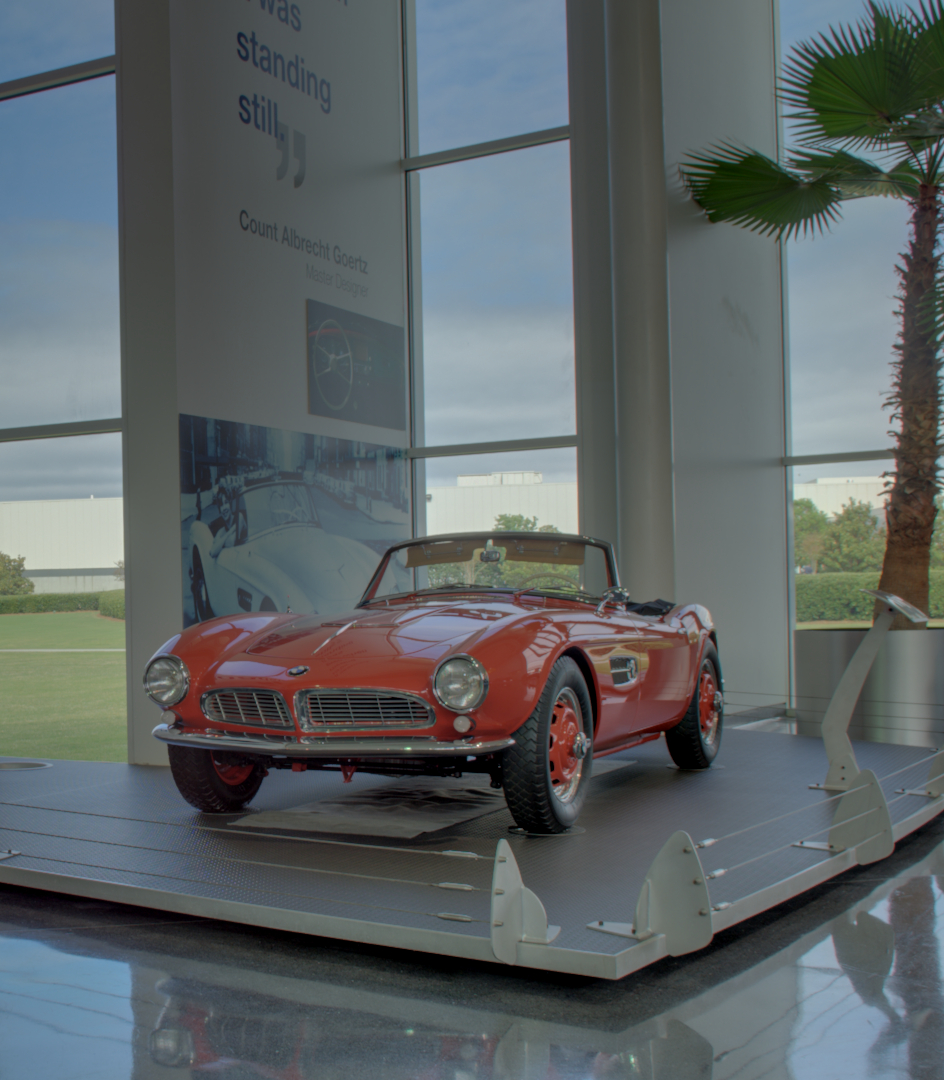}
      \caption{Conventional~\cite{kinoshita2018automatic_trans,kinoshita2019scene}\label{fig:507_conv}}
    \end{subfigure}
    \begin{subfigure}[t]{0.45\hsize}
      \centering
      \includegraphics[width=\columnwidth]{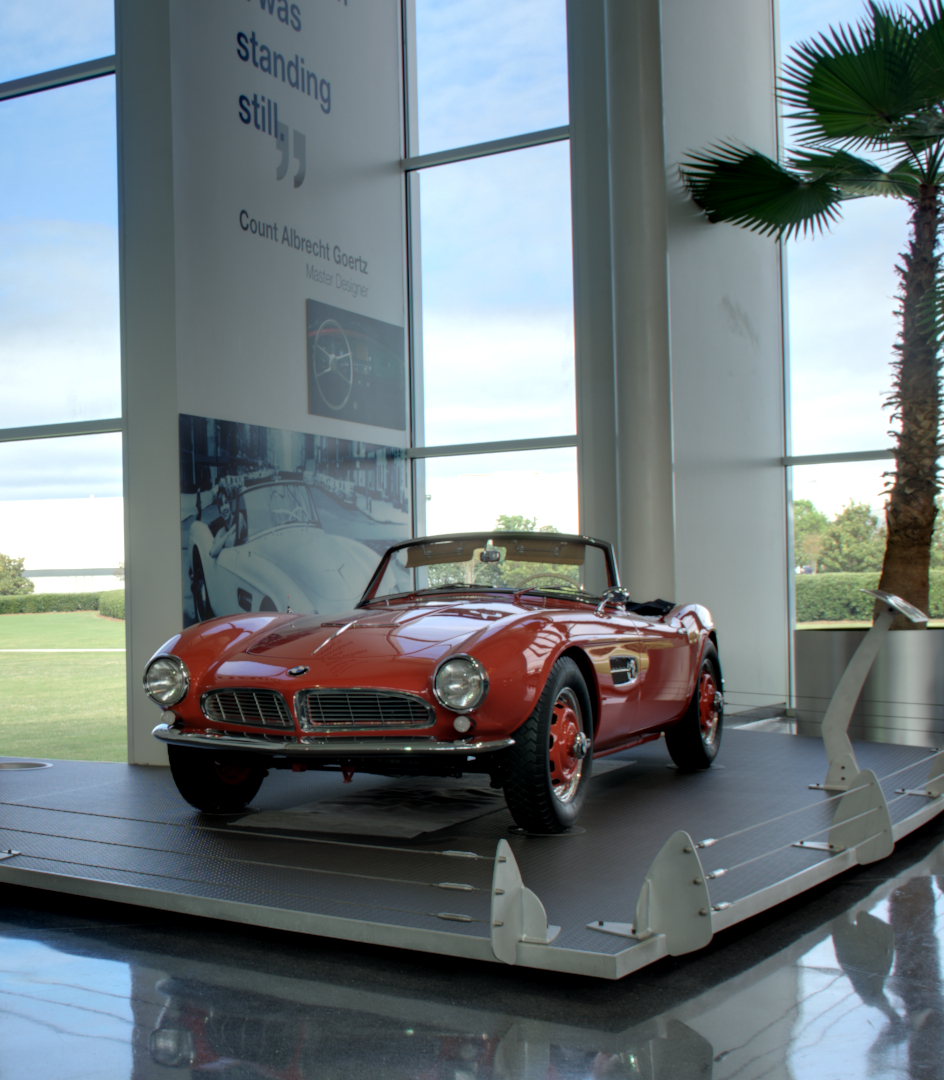}
      \caption{Proposed \label{fig:507_prop}}
    \end{subfigure}
    \caption{Tone-mapped images (``507'') \label{fig:result_507}}
  \end{figure}

\begin{figure}[!t]
    \centering
    \begin{subfigure}[t]{0.45\hsize}
      \centering
      \includegraphics[width=\columnwidth]{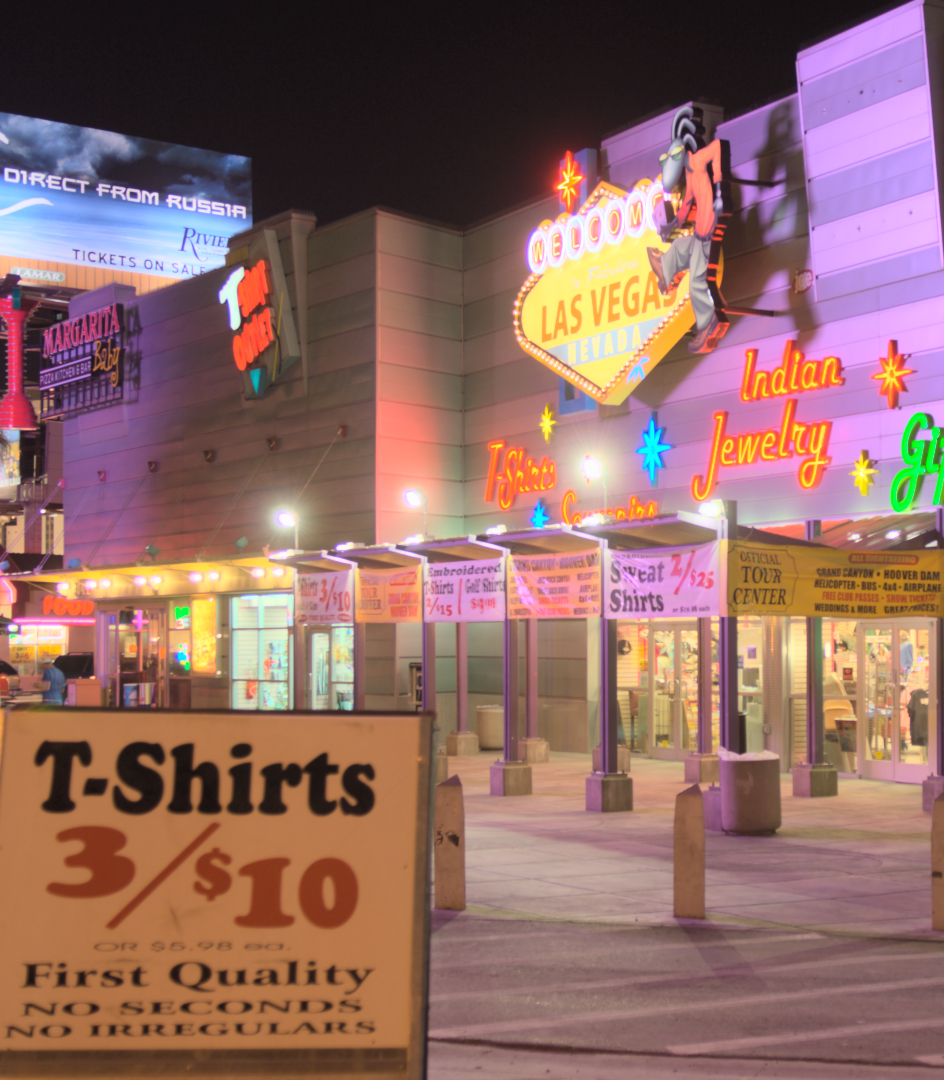}
      \caption{Global~\cite{reinhard2002photographic}\label{fig:lasvegas_global}}
    \end{subfigure}
    \begin{subfigure}[t]{0.45\hsize}
      \centering
      \includegraphics[width=\columnwidth]{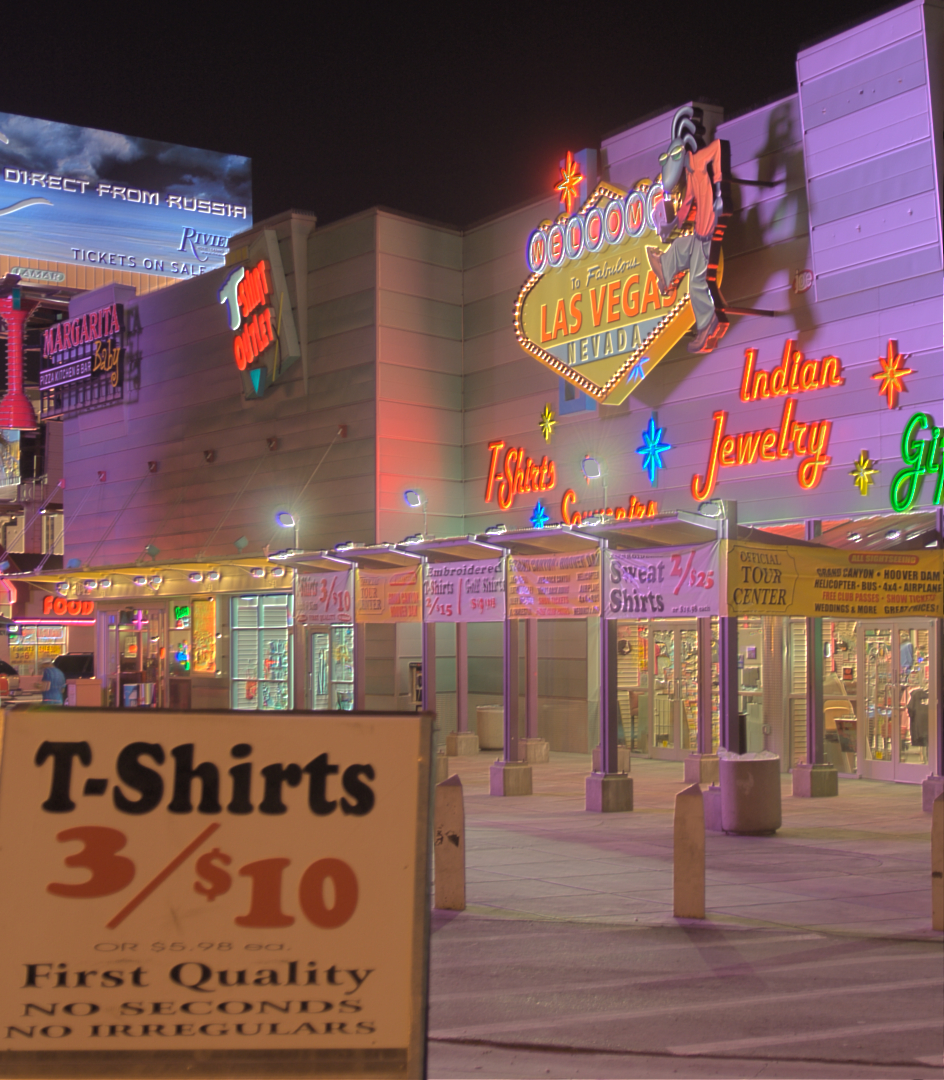}
      \caption{Local~\cite{reinhard2002photographic}\label{fig:lasvegas_local}}
    \end{subfigure}\\
    \begin{subfigure}[t]{0.45\hsize}
      \centering
      \includegraphics[width=\columnwidth]{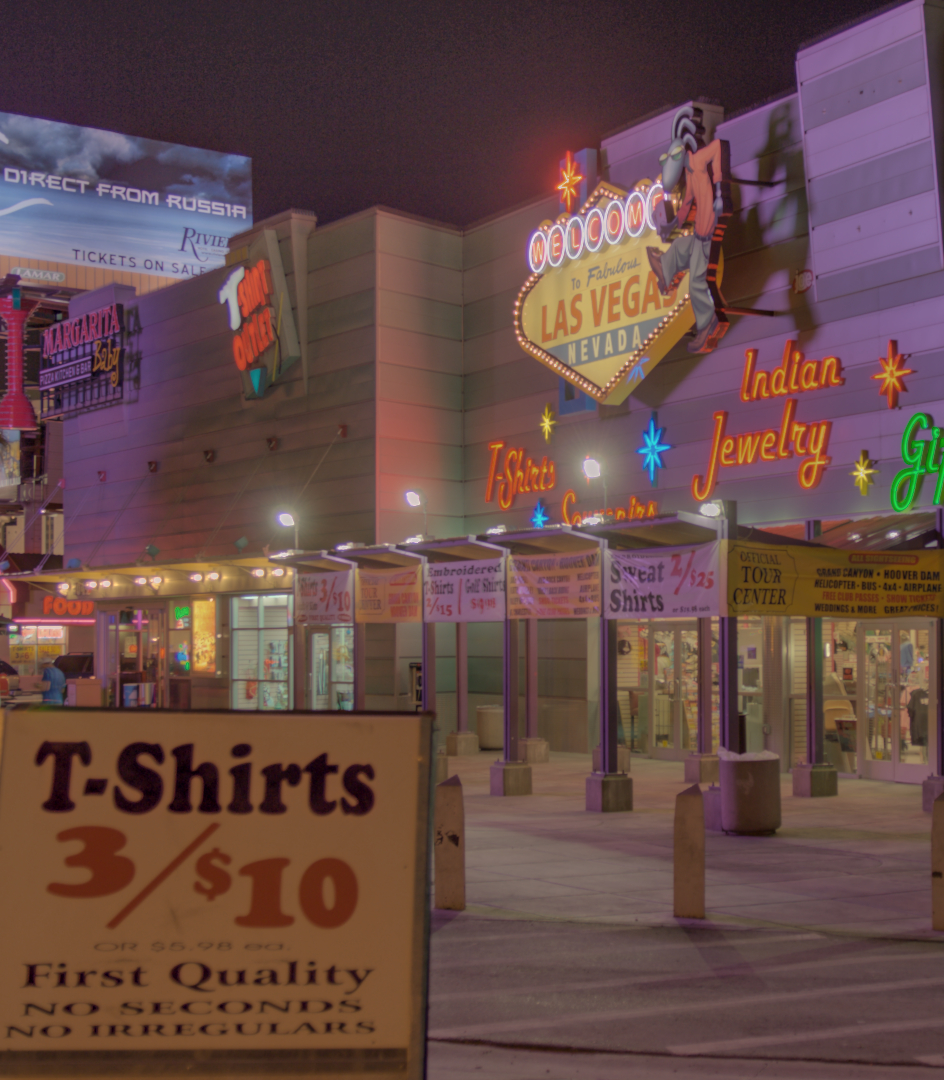}
      \caption{Conventional~\cite{kinoshita2018automatic_trans,kinoshita2019scene}\label{fig:lasvegas_conv}}
    \end{subfigure}
    \begin{subfigure}[t]{0.45\hsize}
      \centering
      \includegraphics[width=\columnwidth]{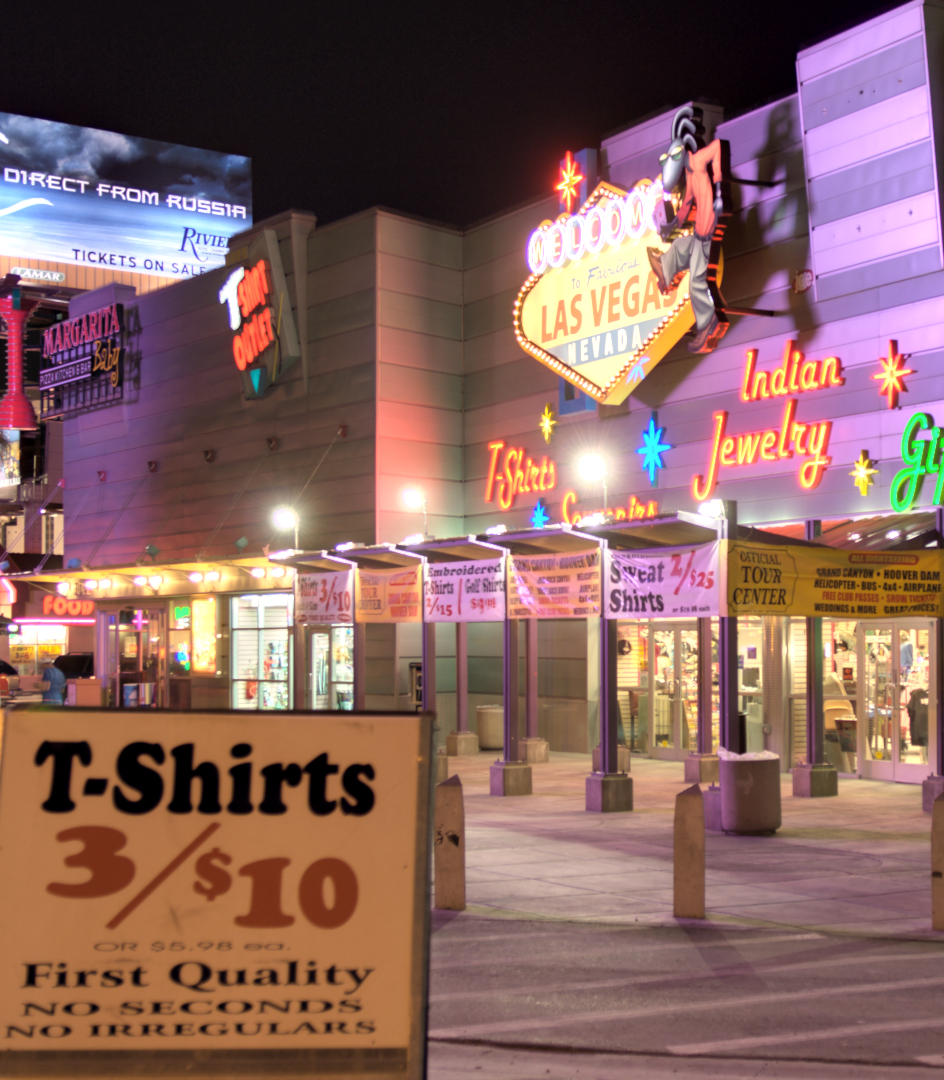}
      \caption{Proposed \label{fig:lasvegas_prop}}
    \end{subfigure}
    \caption{Tone-mapped images (``LasVegasStore'') \label{fig:result_lasvegas}}
  \end{figure}

  Tone-mapped images from input images ``507'' and ``LasVegasStore''
  are shown in \cref{fig:result_507,fig:result_lasvegas}, respectively.
  These figures show that
  the MEF-based conventional method produced grayish, low-contrast images
  have certain effects for enhancement.
  In contrast, the proposed method effectively tone mapped
  HDR images while preserving the contrasts.

  \cref{tab:tmqi} represents TMQI scores of tone mapped images generated by the compared methods
  for four representative scenes
  and their average and standard deviation for all 20 scenes.
  As shown in the table,
  we confirmed that the proposed method provided the highest average of
  TMQI scores.
  This means that the tone-mapped images produced by the proposed method has
  a higher quality than those produced by the other three methods.
  \begin{table}[t!]
    \centering
    \caption{TMQI scores for four representative scenes
      and average and standard deviation for all 20 scenes\label{tab:tmqi}}
    \begin{tabular}{lcccc}\toprule
      Scene               & Global~\cite{reinhard2002photographic} & Local~\cite{reinhard2002photographic}  & Conv.~\cite{kinoshita2018automatic_trans,kinoshita2019scene} & Prop. \\\midrule
      507                 & 0.8737 & 0.8185 & 0.8467       & \textbf{0.9298}   \\
      LabTypewriter       & 0.8323 & 0.7881 & 0.7869       & \textbf{0.8836}   \\
      LasVegasStore       & 0.9204 & 0.8707 & 0.8300       & \textbf{0.9494}   \\
      WillyDesk           & 0.8564 & 0.8077 & 0.8677       & \textbf{0.9326}   \\\midrule
      Average             & 0.8923 & 0.8388 & 0.8628       & \textbf{0.9253}   \\
      Std.                & 0.0398 & 0.0386 & 0.0464       & 0.0395   \\\bottomrule
    \end{tabular}
  \end{table}

\section{Conclusion}
  In this paper, a scene-segmentation-based
  exposure compensation method was proposed
  for MEF based tone mapping of HDR images.
  For the exposure compensation,
  the proposed method segments an input HDR image into subregions
  and determines the exposure value for each subregion
  to maximize the contrast between regions
  while preserving the relative luminance relationships.
  Additionally, to better reflect the effects of exposure compensation
  in the final tone-mapped image,
  a novel fusion weight calculation method based on
  the difference between the luminance and its target
  was also proposed.
  In experiments, MEF-based tone mapping with the proposed compensation method
  and fusion weight calculation outperformed
  typical tone-mapping methods in terms of TMQI.
  Moreover, visual comparison results showed that
  the proposed segmentation-based exposure compensation and fusion weights
  are effective to produce images with high contrast.


\end{document}